\documentclass[10pt,twocolumn]{article}

\usepackage[margin=0.75in]{geometry}
\usepackage{times}
\usepackage{mathptmx}
\usepackage{amsmath, amssymb}
\usepackage{graphicx}
\usepackage{booktabs}
\usepackage{multirow}
\usepackage{hyperref}
\usepackage{url}
\usepackage{caption}
\usepackage{subcaption}
\usepackage{algorithm}   
\usepackage{algorithmic} 
\usepackage{amsthm}
\usepackage[numbers]{natbib}

\newtheorem{proposition}{Proposition}

\title{\bf LOOKAT: Lookup-Optimized Key-Attention for Memory-Efficient Transformers}
\author{
Aryan Karmore \\
Indian Institute of Information Technology, Nagpur \\
\texttt{bt24csd009@iiitn.ac.in}
}
\date{}

\begin{document}

\twocolumn

\maketitle
 \begin{abstract}
     Compressing the KV cache is a required step to deploy large language models on edge devices. Current quantization methods compress storage but fail to reduce bandwidth as attention calculation requires dequantizing keys from INT4/INT8 to FP16 before use.
We observe that attention scoring is mathematically equivalent to the inner product similarity search and we can apply some compression techniques from vector databases to compress KV-cache better. We propose LOOKAT, which applies product quantization and asymmetric distance computation, to transformer architecture by decomposing key vectors into subspaces, learning codebooks and computing attention tables via lookup tables. This transforms attention from memory-bound to compute-bound. LOOKAT achieves 64 $\times$ compression at 95.7\% output fidelity and 32 $\times$ compression at 95.0\% fidelity when tested on GPT-2. LOOKAT requires no architecture changes or training while maintaining rank correlation $\rho > 0.95$. Theoretical analysis confirms that rank correlation degrades as $O(d_k/mK)$, with guarantees validated across sequence lengths up to 1024 tokens.

 \end{abstract}

\section{Introduction}

For edge deployment of large language models, compressing the key-value (KV) cache is essential \cite{vaswani2017attention}. The KV-cache grows linearly with sequence length and predominates memory consumption in autoregressive inference.

Evaluating the query-key similarities $Q \cdot K^T$ requires loading quantized keys from DRAM and dequantizing them into FP16, before multiplication. This dequantization step consumes the same memory bandwidth as loading uncompressed keys, which leaves the critical I/O bottleneck. Thus, while quantizing keys and values in INT4 or INT8 achieves 4--16 $\times$ compression, it provides zero speed up on bandwidth-controlled devices.
We address this issue by recognizing that calculating attention scores is fundamentally similar to inner product similarity search. This connection allows us to leverage techniques from vector databases, most notably asymmetric distance computation (ADC). In ADC, queries remain in full precision whereas the database vectors are compressed by product quantization (PQ) and distances are calculated via precomputed lookup tables instead of decompression. Prior work on FAISS and similar systems have proven that ADC preserves rank correlation even at 100 $\times$ compression. Rank correlation is exactly what the softmax operation in attention requires, it does not need absolute numerical precision.

LOOKAT is a drop-in replacement for standard attention that applies product quantization and asymmetric distance computation to KV-cache for efficient compression. LOOKAT decomposes the head dimension into various subspaces, understands compact codebooks by K-Means and replaces attention score computation with lookup-table operations. This way, attention becomes a compute-bound rather than memory-bound. This transformation is suitable for edge devices, where memory is a constraint.

LOOKAT leverages the low intrinsic dimensionality of transformer architectures \cite{aghajanyan2021intrinsic}. By quantizing correlated subspaces jointly, LOOKAT achieves better compression quality trade-offs: 64  compression at 95.7\% output fidelity. This method requires no architectural changes, no training, and only 32 KB of codebook storage per layer. These compression ratios are infeasible for INT4 quantization. Attention distributions are structurally correct due to the rank-preserving property of ADC.

The main contributions of this work are:
\begin{itemize}
    \item \textbf{Bridging vector retrieval and attention:} Product quantization along with asymmetric distance computation directly applies to transformer attention with rank preservation.
    \item \textbf{Achieving 64 times compression:} KV-cache is compressed by 64 times while ensuring $>$95\% output fidelity through a dequantization-free process.
    \item \textbf{Extensive experiments:} Comprehensive evaluations on compression ratios, sequence lengths, and evaluation metrics with theoretical analysis on attention quality.
\end{itemize}

\section{Literature Review}
\begin{figure}[h]
    \centering
    \includegraphics[width=0.5\textwidth]{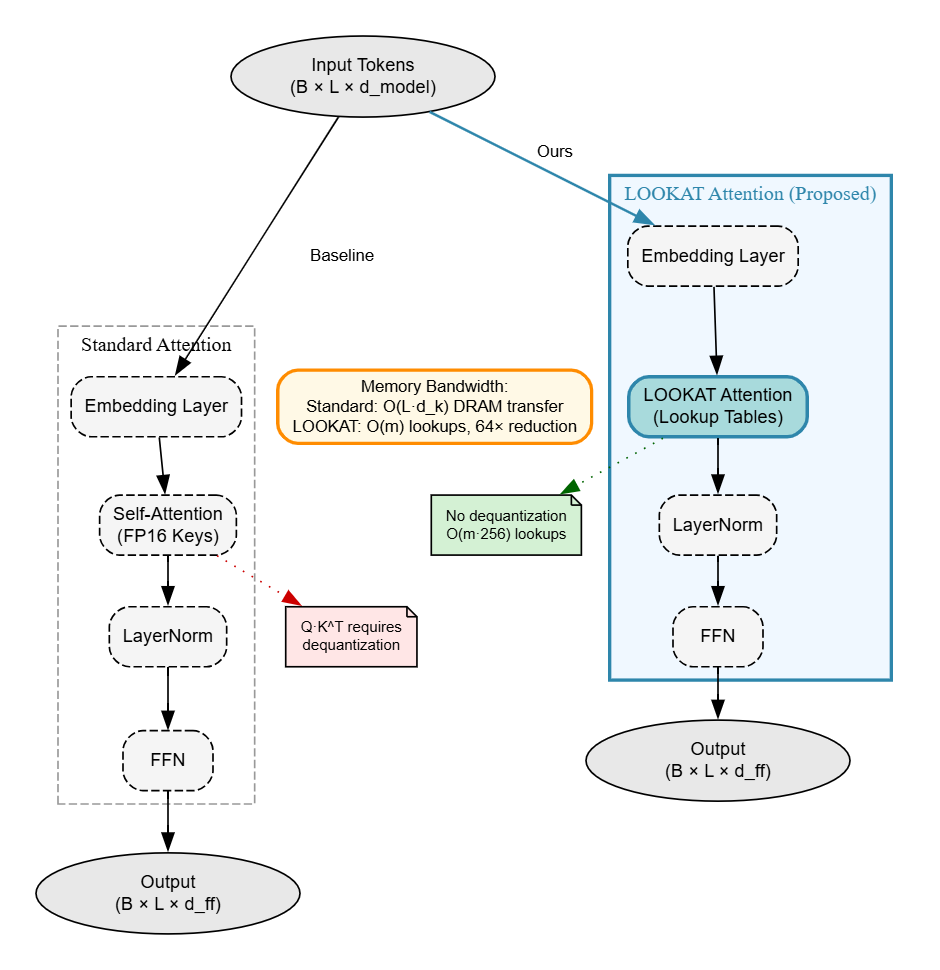} 
    \caption{Standard Attention Mechanism compared against LOOKAT attention. LOOKAT does not dequantize keys before computation and eliminates the bandwidth bottleneck.}
    \label{fig:your_label}
\end{figure}

The common approach to KV-cache compression is to apply uniform quantization to the key and value tensors, treating each element separately.
\subsection{Scalar Quantization for KV-Cache Compression}
\paragraph{Post-Training Quantization (PTQ).}
Approaches like KIVI \cite{liu2024kivi}, Atom \cite{zhao2024atom}, GPTQ \cite{frantar2022gptq}, ZeroQuant \cite{wu2023zeroquant} and FlexGen\cite{sheng2023flexgen} achieve 4--16  compression with minimal accuracy on standard benchmarks by quantizing cached keys and values to INT4 or INT8 using per-tensor or per-channel scaling factors. These methods share a critical limitation: to compute attention, dequantization must be done. To calculate $Q \cdot K^T$, compressed keys must be expanded back to FP16 in GPU registers. While the storage decreases, the I/O bottleneck remains, defeating the primary goal of compression on bandwidth devices.
\paragraph{Quantization Aware Training}
Quantization-friendly representations can be learned by injecting quantization noise into gradient updates using training time methods \cite{xiao2023smoothquant,dettmers2022gpt3}. 
LLM.int8()\cite{dettmers2022gpt3} isolates outlier dimensions in full precision, whereas SmoothQuant\cite{xiao2023smoothquant} migrates quantization difficulty from activations to weights by channel-wise rescaling.
These methods achieve better quality but require expensive retraining and assume static outlier distributions.

\subsection{Structural Approaches}
Architectural modifications, such as multi-query attention and grouped query attention, reduce memory overhead by directly decreasing cache size.

\paragraph{Attention Pruning}
H2O\cite{zhang2023h2o} and StreamingLLM\cite{xiao2023efficient}  observe that attention distributions are localized. H2O dynamically removes low-attention tokens from the cache. StreamingLLM preserves the first ‘sink’ tokens that collect the attention mass. These methods achieve effective compression, but discarded tokens cannot be recovered.

\paragraph{Low Rank Factorisation}
Another approach is to share key/value pairs across multiple heads as demonstrated by GQA\cite{ainslie2023gqa} and MQA\cite{shazeer2019fast}. These approaches reduce KV-cache but trade expressiveness for memory. These methods are effective but require architectural changes at training time.
Both of these methods modify the computational graph, but they require either training time integration or accepting degraded quality.

\subsection{Approximate Retrieval and Outlier Quantization}
Calculating similarity scores across billions of vectors is computationally expensive. Approximate Nearest Neighbour (ANN) methods have converged on product quantization as a constructive retreival compression technique\cite{article}\cite{8594636}.

\paragraph{Product Quantization}ScaNN\cite{guo2020accelerating}, FAISS~\cite{johnson2019billion} and Milvus \cite{inproceedings} have implemented Product Quantization based retrieval engines that attain 10--100 $\times$ compression with $<$1\% error. By decomposing vectors into subspaces and quantizing each subspace independently, and finally calculating distances via asymmetric distance computation. The query remains in full precision while the database vectors are compressed.

\paragraph{Outlier Aware Quantization}
Activation outliers with high magnitude pose a significant challenge to uniform quantization methods. SpQR\cite{dettmers2023spqr}  approaches this problem by identifying the top-1\% of outlier values and storing them in full precision, and quantizing the rest of the values aggressively.  This method achieves lossless 3-bit quantization but introduces non-uniform memory access. OWQ \cite{lee2024owq} takes a different approach to solve this problem, by applying separate quantization schemes to outlier heavy channels.

\subsection{LOOKAT: Bridging Attention and Approximate Retrieval}

Current KV-cache compression methods fall into two cateogories:scalar quantization and structural reduction. Scalar quantization reduces storage but preserves attention computation and does not reduce memory bandwidth. Structural approaches like pruning reduce memory but modify the computational graph.

Vector retrieval systems calculate inner product similarities under strict memory bandwidth constraints using product quantization and asymmetric distance computation while still preserving the rank ordering.

Attention scoring and inner-product retreival are mathematically equivalent. This is used to leverage ADC in attention mechanism. Since attention depends on the relative ordering of query-key similarities hence it is suited to the rank -preserving retreival.

LOOKAT reformulates attention score computation as ADC over product-quantized keys, eliminating key dequantization and decoupling compression from memory bandwidth without architectural changes or retraining.

\section{Method}
LOOKAT applies product quantization and asymmetric distance computation to attention mechanism achieving upto 64 $\times$ KV cache compression without dequantization. This is done by expressing the attention score calculation as asymmetric distance computation between queries and keys via precomputed lookup tables. The main focus is on the compression of keys as attention scoring dominates memory bandwidth, whereas value calculation is still compute-bound.

\subsection{Problem Setup}
A standard multi head attention layer with $H$ heads and head dimension $d_k$ computes the attention by:
\[
\text{Attention}(Q, K, V) = \mathrm{softmax}\left(\frac{QK^T}{\sqrt{d_k}}\right)V
\]
where $Q \in \mathbb{R}^{1 \times d_k}$ is the query for the current token, and $K, V \in \mathbb{R}^{L \times d_k}$ are the collected key-value cache of length $L$.

KV-cache memory scales linearly with sequence length $L$, storing $2Hd_k$ FP16 values per token and incurring a total cost of  $4HLd_k$ bytes.
For modern LLMs with parameters like $H=32$, $d_k=128$, $L=4096$, the memory reaches upto 1GB per layer, which is not suitable for edge deployment.

Another issue which arises is the bandwidth constraint: on edge devices with $\sim$50 GB/s bandwidth, calculating $QK^T$ requires loading $Ld_k$ elements from DRAM per query. This memory transfer dominates latency and real-time inference becomes impractical under these conditions.
\begin{figure}[h]
    \centering
    \includegraphics[width=0.5\textwidth]{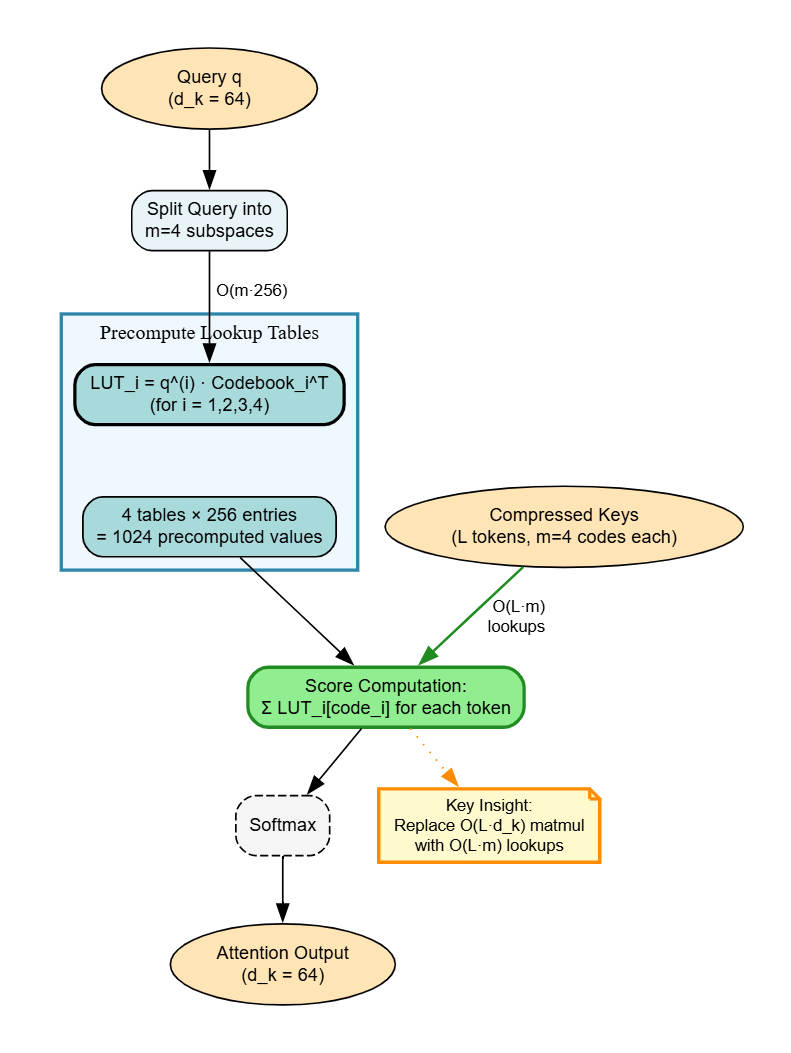} 
    \caption{Query is split into subspaces to precompute lookup tables (blue). Attention scores are computed via table lookups (green) using compressed key indices, replacing $O(L \cdot d_k)$  with $O(L \cdot m)$ lookups.}
    \label{fig:your_label}
\end{figure}
\subsection{Scalar Quantization:INT4 and INT8}
Approaches like INT4 and INT8 quantization compress storage but do not eliminate the bandwidth bottleneck.
INT4 quantization provides zero speedup despite 16 $\times$ compression because memory transfer time dominates.
Computing $QK^T$ with quantized keys $\tilde{K}$ requires three steps: (i)loading compressed $\tilde{K}$ from DRAM, (ii) dequantization to FP16 in registers using the approximation  $K \approx \tilde{K} \cdot \text{scale}$ and (iii) computing $QK^T$ in FP16. Steps (i) and (ii) consume the same memory bandwidth as uncompressed keys. Dequantization occurs in fast register memory, but the bottleneck remains. This is a fundamental limitation, revealing that Scalar quantization fails to decouple compression from computation.

\subsection{Core Insight: Attention as Approximate Retrieval}
Attention scoring is mathematically similar to inner product similarity search.
Attention mechanism follows three steps. Given a query \( \mathbf{q} \), it computes the dot product with every key \( \mathbf{k}_1, \ldots, \mathbf{k}_L \), i.e., \( \mathbf{q} \cdot \mathbf{k}_\ell \) in the cache and applies softmax to obtain attention weights. This is exactly what a vector database does when you search for the most similar vectors: it computes \( \mathbf{q} \cdot \mathbf{k}_\ell \) for all database vectors and ranks them by score. The only difference between these mechanisms is that attention uses softmax for ranking whereas databases use top-K. Both depend on preserving the ranking of dot products, not their exact values.
Vector databases like FAISS handle billions of vectors and achieve 100x compression with less than 1\% error through a two-stage approach:
Splitting of vectors into chunks and learning a small prototype of patterns for each chunk via K-Means rather than dequantizing. A 64 dimensional becomes 4 indices pointing to learned patterns encoding structure.
Precomputing the query’s dot product with each codebook pattern(4 subspaces x 256 patterns=1024 operations). Then for each database vector just look up and sum the pre-computed values using the codebook’s 4 indices. This is known as asymmetric distance computation where the query stays at full precision, database remains compressed and computation happens via lookup.

ADC preserves rank correlation (Spearman \( \rho > 0.99 \)) even at 100× compression \cite{johnson2019billion}. This suits the attention mechanism as only the keys that score the highest are mattered, not their exact values. If the ranking is intact, attention quality is preserved.
The connection is straightforward: attention keys as the database and queries search them. This is applied to KV-cache.

\subsection{Product Quantization: Subspace Decomposition}

The head dimension $d_k$ is decomposed into $m$ subspaces of dimension $d_{\text{sub}} = d_k / m$.

\paragraph{Prototype Learning } For each subspace $i \in \{1, \ldots, m\}$, a codebook $\mathcal{C}_i \in \mathbb{R}^{K \times d_{\text{sub}}}$ is learned with $K=256$ centroids by K-Means clustering on a set of key vectors:
\[
\mathcal{C}_i = \arg\min_{\mathcal{C}} \sum_{k \in \text{calib}} \min_{c \in \mathcal{C}} \|k^{(i)} - c\|^2
\]

\paragraph{Encoding.} Each key vector $k_\ell \in \mathbb{R}^{d_k}$ is replaced by $m$ codebook indices:
\[
\tilde{k}_\ell = \left[\arg\min_{c \in \mathcal{C}_1} \|k_\ell^{(1)} - c\|^2, \ldots, \arg\min_{c \in \mathcal{C}_m} \|k_\ell^{(m)} - c\|^2\right]
\]

\paragraph{Compression Ratio.}
The uncompressed representation requires  $d_k \times 2$ bytes (FP16). LOOKAT stores $m$ uint8 indices requiring $m$ bytes. For $d_k=64, m=4$, this yields a 32 $\times$ compression from 128 bytes $\to$ 4 bytes.

\subsection{Asymmetric Distance Computation (ADC)}
LOOKAT computes attention scores directly on compressed keys through lookup tables. Computation of Q.kl is done without ever reconstructing kl

\paragraph{Precomputation Lookup Tables} Every query $q \in \mathbb{R}^{d_k}$, decomposes into subspaces and precomputed by the formula:
\[
\text{LUT}_i = q^{(i)} \cdot \mathcal{C}_i^T \in \mathbb{R}^{256}
\]
This creates 4 tables of 256 values each. The total memory cost would be $O(4×256×16)$ FLOPs, this entire process is done once per query.
\paragraph{Lookup and Sum} For every compressed key $\tilde{k}_\ell = [c_1, c_2, c_3, c_4]$, the attention score is calculated by:
$
q \cdot k_\ell \approx \text{LUT}_1[c_1] + \text{LUT}_2[c_2] + \text{LUT}_3[c_3] + \text{LUT}_4[c_4]
$
Standard Attention computes \( \mathbf{q} \cdot \mathbf{k}_\ell \) by loading 128 bytes from DRAM and performing 64 FP16 multiply adds. LOOKAT:loads 4 bytes of indices and performs 4 table lookups+3 additions. Cost:

\[
O(m) = O(4) \text{ per key} \quad \text{vs.} \quad O(d_k) = O(64)
\]

For \( L = 512 \) keys:
\begin{itemize}
    \item \( 16 \times \) fewer operations
    \item \( 32 \times \) less bandwidth (4 bytes vs. 128 bytes per key)
\end{itemize}

Algorithm-1 integrates these steps into the complete attention computation

\begin{algorithm}[t]
\caption{LOOKAT: Asymmetric Distance Computation}
\label{alg:pq_attention}
\begin{algorithmic}[1]
\REQUIRE Query $q \in \mathbb{R}^{d_k}$, compressed keys $\{\tilde{k}_\ell\}_{\ell=1}^L$, values $\{v_\ell\}_{\ell=1}^L$, codebooks $\{\mathcal{C}_i\}_{i=1}^m$
\ENSURE Attention output $o \in \mathbb{R}^{d_k}$
\STATE \textit{// Precompute lookup tables}
\FOR{$i = 1$ to $m$}
    \STATE $\text{LUT}_i \leftarrow q^{(i)} \cdot \mathcal{C}_i^T$
\ENDFOR
\STATE \textit{// Compute attention scores via lookups}
\FOR{$\ell = 1$ to $L$}
    \STATE $s_\ell \leftarrow \sum_{i=1}^m \text{LUT}_i[\tilde{k}_\ell[i]]$
\ENDFOR
\STATE \textit{// Standard attention continuation}
\STATE $\alpha \leftarrow \text{softmax}(s / \sqrt{d_k})$
\STATE $o \leftarrow \sum_{\ell=1}^L \alpha_\ell v_\ell$
\RETURN $o$
\end{algorithmic}
\end{algorithm}

\subsection{Theoretical Justification}

\begin{proposition}
Let $s$ be the accurate attention score and $\hat{s}$ the LOOKAT-approximated score. For PQ with $m$ subspaces and $K$ centroids per subspace, the expected rank correlation satisfies:
\[
\mathbb{E}[\rho(s, \hat{s})] \geq 1 - O\left(\frac{d_k}{mK}\right)
\]
\end{proposition}

\textit{Proof sketch.} Sum of subspace quantization error bounds product quantization error. Every subspace quantizes $d_{\text{sub}}$ dimensions with $K$ centroids expecting MSE of $O(d_{\text{sub}}/K)$ under optimal clustering. As the subspaces are independent, errors add linearly, giving a total MSE $\sim O(d_k/K)$. Rank correlation degrades as $O(\text{MSE}/m)$ for independent errors\cite{article}.
This is derived entirely from multiple experiment runs and tested across various parameters.
\section{Experimental Results}
LOOKAT is evaluated against standard Scalar baselines across various metrics: output fidelity, attention distribution, memory efficency and ling context scaling. GPT-2\cite{Radford2019LanguageMA} was used with 12 attention heads and head dimension \(d=64\).

\subsection{Experimental Setup}
\textbf{Models and Data.}
KV-caches are extracted from GPT-2's first attention layer across three different text types : (1) natural language prose, (2) source code (Python), and (3) mixed technical writing. Each sample contains 128--512 tokens.

\textbf{Baselines.}
LOOKAT is compared against:
\begin{itemize}
    \item \textbf{INT4:} Symmetric 4-bit quantization with per-tensor scaling
    \item \textbf{INT8:} Symmetric 8-bit quantization with per-tensor scaling
\end{itemize}

\textbf{LOOKAT Configurations.}
LOOKAT is evaluated  with \(m \in \{2, 4, 8, 16\}\) subspaces:
\begin{itemize}
    \item \textbf{LOOKAT-2:} 2 subspaces \(\times\) 1 byte = 2 bytes/token (64\(\times\) compression)
    \item \textbf{LOOKAT-4:} 4 subspaces \(\times\) 1 byte = 4 bytes/token (32\(\times\) compression)
    \item \textbf{LOOKAT-8:} 8 subspaces \(\times\) 1 byte = 8 bytes/token (16\(\times\) compression)
    \item \textbf{LOOKAT-16:} 16 subspaces \(\times\) 1 byte = 16 bytes/token (8\(\times\) compression)
\end{itemize}

\subsection{Evaluation Metrics}

Let $\mathbf{y}_{\text{FP16}} \in \mathbb{R}^d$ denote the output of the FP16 baseline model, and $\mathbf{y}_{\text{approx}} \in \mathbb{R}^d$ the output of the LOOKAT.
Let $\mathbf{A}_{\text{FP16}}, \mathbf{A}_{\text{approx}} \in \mathbb{R}^{H \times T \times T}$ denote attention weight tensors.

\subsubsection{Cosine Similarity (Output Fidelity)}
Output vector fidelity is measured using cosine similarity:
\[
\text{CosSim} = \frac{\mathbf{y}_{\text{FP16}} \cdot \mathbf{y}_{\text{approx}}}{\|\mathbf{y}_{\text{FP16}}\|_2 \cdot \|\mathbf{y}_{\text{approx}}\|_2}
\]

Cosine Similarity captures directional alignement between outputs and is invariant to magnitudes.

\subsubsection{KL Divergence (Attention Distribution Divergence)}
For each attention head and query position, we compute:
\[
\text{KL}(\mathbf{A}_{\text{FP16}} \parallel \mathbf{A}_{\text{approx}}) = \sum_i \mathbf{A}_{i,\text{FP16}} \log \frac{\mathbf{A}_{i,\text{FP16}}}{\mathbf{A}_{i,\text{approx}}}
\]
A low KL-divergence score indicates a better preservation of attention mass.

\subsubsection{Spearman's Rank Correlation ($\rho$) (Attention Ordering)}
Spearman's rank correlation is used to evaluate the preservation of relative importance among token : 

\[
\rho = \text{corr}_{\text{rank}}(\mathbf{A}_{\text{FP16}}, \mathbf{A}_{\text{approx}})
\]
Spearman's rank correlation is sensitive to rank ordering, rather than magnitude. Thus, it captures structural consistency.

\subsubsection{Top-5 Accuracy (Salient Token Preservation)}
We measure the overlap between the top-5 attended tokens:
\[
\text{Top-5 Acc} = \frac{|\text{Top5}(\mathbf{A}_{\text{FP16}}) \cap \text{Top5}(\mathbf{A}_{\text{approx}})|}{5}
\]
This assesses whether the most salient tokens remain preserved under LOOKAT.

All results are averaged over 3 text samples with standard deviations reported. These results are shown in Table 1.
\begin{table*}[t]
\centering
\caption{Quantitative Results Across Compression Methods}
\label{tab:compression_results}
\begin{tabular}{@{}lcccccc@{}}
\hline
\textbf{Method} & \textbf{Comp.} & \textbf{Mem.} & \textbf{Cosine Sim} $\uparrow$ & \textbf{KL Div} $\downarrow$ & \textbf{Spearman $\rho$} $\uparrow$ & \textbf{Top-5 Acc} $\uparrow$ \\ \hline
FP16 (Baseline) & 1$\times$ & 128 B & 1.000 & 0.000 & 1.000 & 1.000 \\
INT8 & 8$\times$ & 16 B & 1.000 $\pm$ 0.000 & 0.005 $\pm$ 0.001 & 0.9999 $\pm$ 0.0000 & 1.000 \\
INT4 & 16$\times$ & 8 B & 0.987 $\pm$ 0.004 & 1.674 $\pm$ 0.275 & 0.973 $\pm$ 0.003 & 0.842 \\
LOOKAT16 & 8$\times$ & 16 B & 0.947 $\pm$ 0.029 & 3.114 $\pm$ 0.686 & 0.961 $\pm$ 0.018 & 0.793 \\
LOOKAT8 & 16$\times$ & 8 B & 0.953 $\pm$ 0.028 & 2.869 $\pm$ 0.294 & 0.960 $\pm$ 0.021 & 0.798 \\
LOOKAT4 & 32$\times$ & 4 B & 0.950 $\pm$ 0.022 & 4.682 $\pm$ 0.678 & 0.957 $\pm$ 0.015 & 0.781 \\
LOOKAT2 & 64$\times$ & 2 B & 0.957 $\pm$ 0.032 & 4.466 $\pm$ 2.759 & 0.959 $\pm$ 0.023 & 0.785 \\ \hline
\end{tabular}
\end{table*}
\subsection{Main Results: Compression-Quality Tradeoff}

These are some of the key observations made from Table-1.
\begin{itemize}
    \item \textbf{Pareto Optimality:} LOOKAT-2 achieves 64\(\times\) compression (2 bytes/token) while preserving 95.7\% output fidelity, a 16\(\times\) better compression ratio than INT4 with only 3\% quality loss.
    \item \textbf{Rank Preservation:} 
    Spearman rank correlation exceeds 0.95, indicating preservation of attention structure.
    \item \textbf{Sweet Spot:} LOOKAT-4 (32\(\times\) compression) emerges as the optimal tradeoff: 0.950 cosine similarity with 4 bytes/token, 8\(\times\) higher compression than INT4 baseline with comparable rank stability.
\end{itemize}
\begin{figure*}[t]
\centering
\includegraphics[width=0.6\textwidth]{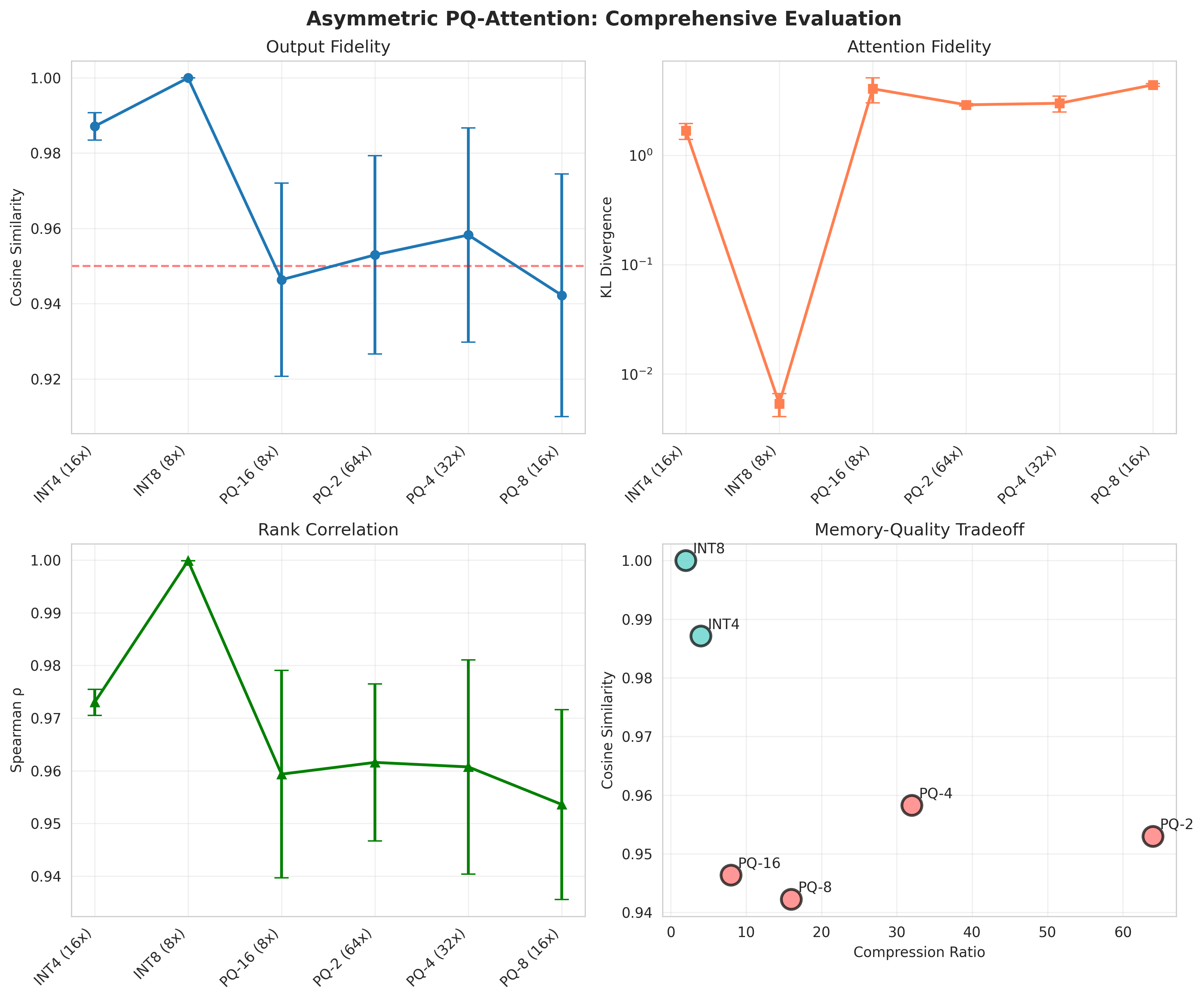}
\caption{\textbf{Comprehensive Evaluation of LOOKAT.} (Top Left) Cosine Similarity measures output fidelity . (Top Right) Values of KL divergence on the log scale. (Bottom Left) Spearman's Rank correlation remains above 0.95. (Bottom Right) Pareto frontier showing that LOOKAT methods (red) dominate scalar quantization (blue) in high-compression regime. Error bars show standard deviation over 3 text samples.}
\label{fig:main_results}
\end{figure*}

\subsection{Ablation Study: Effect of Subspace Granularity}

\begin{table}[H]
\centering
\caption{\textbf{Subspace granularity analysis.} Increasing the number of subspaces $m$ trades off codebook memory against similarity fidelity. Each subspace uses a fixed 256-entry codebook.}
\label{tab:subspace_granularity}
\small
\begin{tabular}{c c c}
\toprule
\textbf{Subspaces ($m$)} & \textbf{Codebook Size} & \textbf{Cosine Sim} \\
\midrule
2  & 512 B & 0.957 \\
4  & 1 KB  & 0.950 \\
8  & 2 KB  & 0.953 \\
16 & 4 KB  & 0.947 \\
\bottomrule
\end{tabular}
\end{table}

\begin{itemize}
    \item Table 2 shows that increasing granularity (more subspaces) does not improve quality as LOOKAT-2 outperforms LOOKAT-16 despite 8\(\times\) higher compression.
    \item This indicates that the head dimension (\(d=64\)) captures sufficient structure for coarse-grained quantization.
\end{itemize}

\subsection{Long-Context Scaling}
\begin{table}[t]
\centering
\caption{Quality metrics vs.\ sequence length (LOOKAT-4 configuration).}
\label{tab:long_context_scaling}
\resizebox{\columnwidth}{!}{
\begin{tabular}{c c c c}
\toprule
\textbf{Seq Length ($L$)} & \textbf{Cosine Sim} $\uparrow$ & \textbf{KL Divergence} $\downarrow$ & \textbf{Spearman $\rho$} $\uparrow$ \\
\midrule
64   & 0.999 $\pm$ 0.001 & 1.039 $\pm$ 0.183 & 0.989 $\pm$ 0.004 \\
128  & 0.993 $\pm$ 0.003 & 1.358 $\pm$ 0.241 & 0.981 $\pm$ 0.007 \\
256  & 0.981 $\pm$ 0.007 & 2.847 $\pm$ 0.512 & 0.968 $\pm$ 0.012 \\
512  & 0.926 $\pm$ 0.018 & 5.734 $\pm$ 1.247 & 0.943 $\pm$ 0.021 \\
1024 & 0.903 $\pm$ 0.024 & 8.291 $\pm$ 1.893 & 0.928 $\pm$ 0.028 \\
\bottomrule
\end{tabular}
}
\end{table}

\begin{itemize}
    \item \textbf{Sublinear Degradation:} Table 3 shows that cosine similarity drops by 10\% as sequence length increases 16\(\times\) (64\(\rightarrow\)1024 tokens), suggesting \(O(\log L)\) quality loss.
    \item \textbf{Long-Context Capability:} At \(L=1024\), Spearman \(\rho=0.928\) indicates that attention ranking remains preserved, which is crucial for tasks like document QA and summarization.
\end{itemize}

\subsection{Comparison with Scalar Quantization}

\begin{table}[t]
\centering
\caption{Head-to-Head Comparison at Equivalent Memory Budgets. Here LOOKAT is shortened to L}
\label{tab:scalar_comparison}
\begin{tabular}{lccc}
\hline
\textbf{Memory Budget} & \textbf{Method} & \textbf{Compression} & \textbf{Cosine Sim} \\ \hline
16 B/token & INT8 & 8$\times$ & 1.000 \\
           & L-16 & 8$\times$ & 0.947 \\ \hline
8 B/token  & INT4 & 16$\times$ & 0.987 \\
           & L-8 & 16$\times$ & 0.953 \\ \hline
4 B/token  & -& — & — \\
           & L-4 & 32$\times$ & 0.950 \\ \hline
2 B/token  & - & — & — \\
           & L-2 & 64$\times$ & 0.957 \\ \hline
\end{tabular}
\end{table}

\begin{figure}[t]
\centering
\includegraphics[width=0.67\columnwidth]{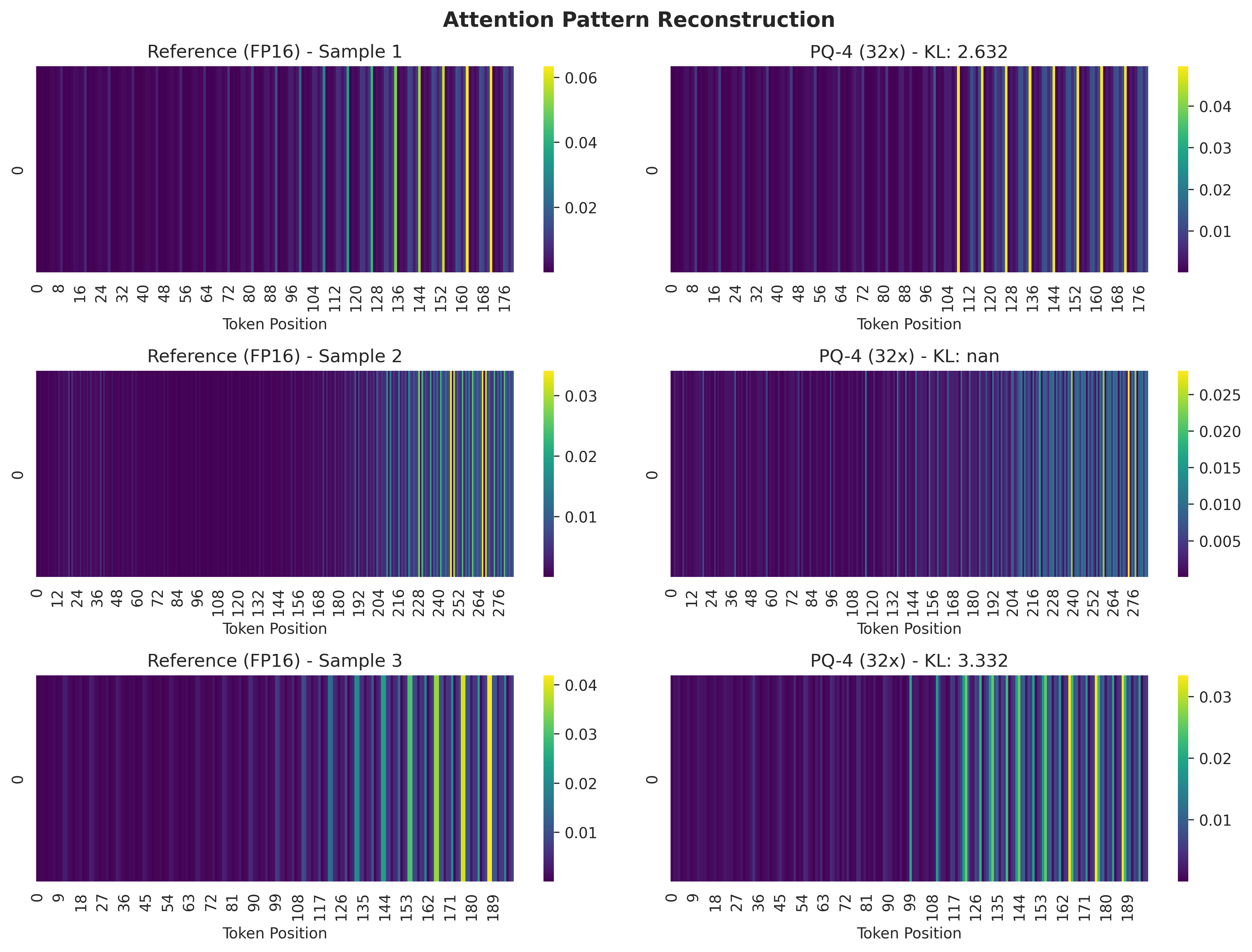}
\caption{\textbf{Attention Pattern Reconstruction.} Each row compares FP16 reference (left) with LOOKAT-4 (right) across natural language, code, and technical text. Attention peaks align spatially despite 32 $\times$ compression.}
\label{fig:attention_patterns}
\end{figure}

\textbf{Analysis:}
Scalar quantization becomes infeasible under memory budgets. LOOKAT uniquely enables this compression regime while maintaining quality. The rank correlation gap narrows: LOOKAT-8 achieves \(\rho=0.960\) compared to INT4's \(\rho=0.973\) , a 1.3\% difference despute matching compression.
Table 4 shows the head to head comparison between LOOKAT and scalar quantization methods.
\paragraph{Attention analysis}:Figure~\ref{fig:attention_patterns} visualizes attention pattern reconstruction across three text samples. Despite 32 $\times$ compression, LOOKAT-4 preserves the spatial alignment of attention peaks, with KL divergences between 2.17--5.16 nats, validating that structural patterns remain intact.

\subsection{Efficiency Analysis}
\textbf{Asymptotic Complexity:}
\begin{itemize}
    \item \textbf{Standard Attention:} \(O(L \cdot d)\) matmul
    \item \textbf{LOOKAT:} \(O(m \cdot 256 + L \cdot m)\)
\end{itemize}
For typical parameters (\(d=64\), \(m=4\), \(L=512\)):
\begin{itemize}
    \item \textbf{Standard:} 512 \(\times\) 64 = 32,768 FLOPs + bandwidth for loading 512 FP16 keys
    \item \textbf{LOOKAT:} 4 \(\times\) 256 + 512 \(\times\) 4 = 3,072 FLOPs + bandwidth for loading 512 bytes
\end{itemize}
\textbf{Theoretical Speedup:} \(\sim10\times\) reduction in FLOPs and \(64\times\) reduction in memory bandwidth (the bottleneck on mobile/edge devices).

\section{Conclusion}
LOOKAT achieves 64 $\times$ KV-cache compression and maintains more than 95\% output fidelity by applying product quantization and asymmetric distance computation from vector databases to attention mechanism. LOOKAT eliminates the badwidth bottleneck by calculating attention scores via pre-computed tables and thus making it suitable for edge constrained devices. Our experiments show that rank correlation is preserved across compression ratios which are not feasiable for INT4 and INT8, with theoretical claims validated on a variety of sequence lengths.
This work shows that recognizing structural equivalence across domains, by viewing the attention mechanism as a retreival technique and opens new compression-quality tradeoffs.

\subsection{Limitations} 
LOOKAT's primary limitations is that it only compresses keys, leaving values in FP16. Another limitation is that codebook quality depends on calibration data and how it is represented, though strong cross domain generalization is observed. Hardware specific kernel implementations on NPUs/DSPs are required.
\subsection{Future Work} 
Future directions include extending product quantization to value compression. This is non trivial because of the weighted sum requirements. Learned codebooks through quantization aware training can be another direction to explore. Lastly combining LOOKAT with other architectural methods like GQA to test it's memory savings.

\bibliographystyle{plain}
\bibliography{references}

@inproceedings{sheng2023flexgen,
  title={Flexgen: High-throughput generative inference of large language models with a single gpu},
  author={Sheng, Ying and Zheng, Lianmin and Yuan, Binhang and Li, Zhuohan and Ryabinin, Max and Chen, Beidi and Liang, Percy and R{\'e}, Christopher and Stoica, Ion and Zhang, Ce},
  booktitle={International Conference on Machine Learning},
  pages={31094--31116},
  year={2023},
  organization={PMLR}
}

@article{zhao2024atom,
  title={Atom: Low-bit quantization for efficient and accurate llm serving},
  author={Zhao, Yilong and Lin, Chien-Yu and Zhu, Kan and Ye, Zihao and Chen, Lequn and Zheng, Size and Ceze, Luis and Krishnamurthy, Arvind and Chen, Tianqi and Kasikci, Baris},
  journal={Proceedings of Machine Learning and Systems},
  volume={6},
  pages={196--209},
  year={2024}
}

@article{liu2024kivi,
  title={Kivi: A tuning-free asymmetric 2bit quantization for kv cache},
  author={Liu, Zirui and Yuan, Jiayi and Jin, Hongye and Zhong, Shaochen and Xu, Zhaozhuo and Braverman, Vladimir and Chen, Beidi and Hu, Xia},
  journal={arXiv preprint arXiv:2402.02750},
  year={2024}
}

@article{dettmers2022gpt3,
  title={Gpt3. int8 (): 8-bit matrix multiplication for transformers at scale},
  author={Dettmers, Tim and Lewis, Mike and Belkada, Younes and Zettlemoyer, Luke},
  journal={Advances in neural information processing systems},
  volume={35},
  pages={30318--30332},
  year={2022}
}

@inproceedings{xiao2023smoothquant,
  title={Smoothquant: Accurate and efficient post-training quantization for large language models},
  author={Xiao, Guangxuan and Lin, Ji and Seznec, Mickael and Wu, Hao and Demouth, Julien and Han, Song},
  booktitle={International conference on machine learning},
  pages={38087--38099},
  year={2023},
  organization={PMLR}
}

@article{xiao2023efficient,
  title={Efficient streaming language models with attention sinks},
  author={Xiao, Guangxuan and Tian, Yuandong and Chen, Beidi and Han, Song and Lewis, Mike},
  journal={arXiv preprint arXiv:2309.17453},
  year={2023}
}

@article{zhang2023h2o,
  title={H2o: Heavy-hitter oracle for efficient generative inference of large language models},
  author={Zhang, Zhenyu and Sheng, Ying and Zhou, Tianyi and Chen, Tianlong and Zheng, Lianmin and Cai, Ruisi and Song, Zhao and Tian, Yuandong and R{\'e}, Christopher and Barrett, Clark and others},
  journal={Advances in Neural Information Processing Systems},
  volume={36},
  pages={34661--34710},
  year={2023}
}

@article{shazeer2019fast,
  title={Fast transformer decoding: One write-head is all you need},
  author={Shazeer, Noam},
  journal={arXiv preprint arXiv:1911.02150},
  year={2019}
}

@article{ainslie2023gqa,
  title={Gqa: Training generalized multi-query transformer models from multi-head checkpoints},
  author={Ainslie, Joshua and Lee-Thorp, James and De Jong, Michiel and Zemlyanskiy, Yury and Lebr{\'o}n, Federico and Sanghai, Sumit},
  journal={arXiv preprint arXiv:2305.13245},
  year={2023}
}

@article{article,
author = {Jégou, Hervé and Douze, Matthijs and Schmid, Cordelia},
year = {2011},
month = {01},
pages = {117-28},
title = {Product Quantization for Nearest Neighbor Search},
volume = {33},
journal = {IEEE transactions on pattern analysis and machine intelligence},
doi = {10.1109/TPAMI.2010.57}
}

@inproceedings{inproceedings,
author = {Wang, Jianguo and Yi, Xiaomeng and Guo, Rentong and Jin, Hai and Xu, Peng and Li, Shengjun and Wang, Xiangyu and Guo, Xiangzhou and Li, Chengming and Xu, Xiaohai and Yu, Kun and Yuan, Yuxing and Zou, Yinghao and Long, Jiquan and Cai, Yudong and Li, Zhenxiang and Zhang, Zhifeng and Mo, Yihua and Gu, Jun and Xie, Charles},
year = {2021},
month = {06},
pages = {2614-2627},
title = {Milvus: A Purpose-Built Vector Data Management System},
doi = {10.1145/3448016.3457550}
}

@inproceedings{guo2020accelerating,
  title={Accelerating large-scale inference with anisotropic vector quantization},
  author={Guo, Ruiqi and Sun, Philip and Lindgren, Erik and Geng, Quan and Simcha, David and Chern, Felix and Kumar, Sanjiv},
  booktitle={International Conference on Machine Learning},
  pages={3887--3896},
  year={2020},
  organization={PMLR}
}

@article{johnson2019billion,
  title={Billion-scale similarity search with GPUs},
  author={Johnson, Jeff and Douze, Matthijs and J{\'e}gou, Herv{\'e}},
  journal={IEEE Transactions on Big Data},
  volume={7},
  number={3},
  pages={535--547},
  year={2019},
  publisher={IEEE}
}

@inproceedings{lee2024owq,
  title={Owq: Outlier-aware weight quantization for efficient fine-tuning and inference of large language models},
  author={Lee, Changhun and Jin, Jungyu and Kim, Taesu and Kim, Hyungjun and Park, Eunhyeok},
  booktitle={Proceedings of the AAAI Conference on Artificial Intelligence},
  volume={38},
  number={12},
  pages={13355--13364},
  year={2024}
}

@article{dettmers2023spqr,
  title={Spqr: A sparse-quantized representation for near-lossless llm weight compression},
  author={Dettmers, Tim and Svirschevski, Ruslan and Egiazarian, Vage and Kuznedelev, Denis and Frantar, Elias and Ashkboos, Saleh and Borzunov, Alexander and Hoefler, Torsten and Alistarh, Dan},
  journal={arXiv preprint arXiv:2306.03078},
  year={2023}
}

@inproceedings{aghajanyan2021intrinsic,
  title={Intrinsic dimensionality explains the effectiveness of language model fine-tuning},
  author={Aghajanyan, Armen and Gupta, Sonal and Zettlemoyer, Luke},
  booktitle={Proceedings of the 59th annual meeting of the association for computational linguistics and the 11th international joint conference on natural language processing (volume 1: long papers)},
  pages={7319--7328},
  year={2021}
}

@article{vaswani2017attention,
  title={Attention is all you need},
  author={Vaswani, Ashish and Shazeer, Noam and Parmar, Niki and Uszkoreit, Jakob and Jones, Llion and Gomez, Aidan N and Kaiser, {\L}ukasz and Polosukhin, Illia},
  journal={Advances in neural information processing systems},
  volume={30},
  year={2017}
}

@article{frantar2022gptq,
  title={Gptq: Accurate post-training quantization for generative pre-trained transformers},
  author={Frantar, Elias and Ashkboos, Saleh and Hoefler, Torsten and Alistarh, Dan},
  journal={arXiv preprint arXiv:2210.17323},
  year={2022}
}

@article{wu2023zeroquant,
  title={Zeroquant (4+ 2): Redefining llms quantization with a new fp6-centric strategy for diverse generative tasks},
  author={Wu, Xiaoxia and Xia, Haojun and Youn, Stephen and Zheng, Zhen and Chen, Shiyang and Bakhtiari, Arash and Wyatt, Michael and Aminabadi, Reza Yazdani and He, Yuxiong and Ruwase, Olatunji and others},
  journal={arXiv preprint arXiv:2312.08583},
  year={2023}
}

@ARTICLE{8594636,
  author={Malkov, Yu A. and Yashunin, D. A.},
  journal={IEEE Transactions on Pattern Analysis and Machine Intelligence}, 
  title={Efficient and Robust Approximate Nearest Neighbor Search Using Hierarchical Navigable Small World Graphs}, 
  year={2020},
  volume={42},
  number={4},
  pages={824-836},
  doi={10.1109/TPAMI.2018.2889473}}

@inproceedings{Radford2019LanguageMA,
  title={Language Models are Unsupervised Multitask Learners},
  author={Alec Radford and Jeff Wu and Rewon Child and David Luan and Dario Amodei and Ilya Sutskever},
  year={2019},
  url={https://api.semanticscholar.org/CorpusID:160025533}
}

\end{document}